\makeatletter\def\graphicscache@inhibit{true}\makeatother

\documentclass[runningheads]{llncs}

\usepackage[utf8]{inputenc}
\usepackage{lmodern}
\usepackage[T1]{fontenc}
\usepackage{microtype}

\usepackage{graphicx}
\usepackage{comment}
\usepackage{amsmath,amssymb}
\usepackage{color}
\usepackage{url}
\usepackage{hyperref}
\usepackage{tabularx}
\usepackage{threeparttable}
\usepackage[binary-units=true,product-units=single,per-mode=symbol,range-units=single,range-phrase=\,--\,,detect-all]{siunitx}
\DeclareSIUnit\pixel{px}

\usepackage{tikz}
\usepackage{tkz-graph}
\usepackage{tikz-3dplot}
\usepackage{pgfplots}
\usetikzlibrary{arrows}
\usetikzlibrary{positioning,calc}
\usetikzlibrary{decorations.pathreplacing}
\usetikzlibrary{decorations.markings}
\usetikzlibrary{fit}
\usetikzlibrary{shapes.callouts}
\usetikzlibrary{shapes.geometric}
\usetikzlibrary{shapes.arrows}
\usetikzlibrary{matrix}
\usetikzlibrary{fit,shapes.callouts,shapes.geometric,backgrounds,positioning,arrows.meta,calc}
\usepackage{pgfplots}
\pgfplotsset{compat=1.9}
\usepgfplotslibrary{groupplots}
\pgfplotsset{every axis/.append style={label style={font=\small},tick label style={font=\small},},}
\usepackage{calc}
\usepackage{orcidlink}

\renewcommand\vec{\mathbf}

\newcommand{\reffig}[1]{Fig.~\ref{#1}}
\newcommand{\reftab}[1]{Tab.~\ref{#1}}

\newcommand{\etal}{et al.~}

\newcommand{\dotss}{\mathrel{{.}\,{.}}\nobreak}

\graphicspath{{./figures/}}

\newif\ifreview
\reviewfalse

\ifreview
	\usepackage{lineno}

	\linenumbers
\fi

\begin{document}

\def\SubNumber{29}

\def\GCPRTrack{Young Researcher's Forum}
\title{Online Marker-free Extrinsic Camera Calibration using Person Keypoint Detections
\thanks{This work was funded by grant BE 2556/16-2 (Research Unit FOR 2535 Anticipating Human Behavior) of the German Research Foundation (DFG).}}
\renewcommand\footnotemark{}

\ifreview
\else
	\titlerunning{Online Marker-free Extr. Camera Calib. using Person Keypoint Detections}

	\author{Bastian Pätzold\inst{}\orcidlink{0000-0002-6395-5854} \and
	Simon Bultmann\inst{}\orcidlink{0000-0001-9509-2080} \and
	Sven Behnke\inst{}\orcidlink{0000-0002-5040-7525}}
	
	\authorrunning{B. Pätzold et al.}

	\institute{University of Bonn, Institute for Computer Science VI, Autonomous Intelligent Systems, Friedrich-Hirzebruch-Allee 8, 53115 Bonn, Germany\\
	\email{\{paetzold,bultmann,behnke\}@ais.uni-bonn.de}\\
	\url{https://www.ais.uni-bonn.de}}
	
\fi

\maketitle              %

\begin{tikzpicture}[remember picture,overlay]
	\node[anchor=north,align=left,font=\sffamily,yshift=-0.2cm] at (current page.north) {%
	  DAGM German Conference on Pattern Recognition (GCPR), Konstanz, September 2022.
	};
\end{tikzpicture}%
	
\begin{abstract}
Calibration of multi-camera systems, i.e. determining the relative poses between the cameras, is a prerequisite for many tasks in computer vision and robotics. Camera calibration is typically achieved using offline methods that use checkerboard calibration targets. These methods, however, often are cumbersome and lengthy, considering that a new calibration is required each time any camera pose changes.
In this work, we propose a novel, marker-free online method for the extrinsic calibration of multiple smart edge sensors, relying solely on 2D human keypoint detections that are computed locally on the sensor boards from RGB camera images. Our method assumes the intrinsic camera parameters to be known and requires priming with a rough initial estimate of the camera poses.
The person keypoint detections from multiple views are received at a central backend where they are synchronized, filtered, and assigned to person hypotheses.
We use these person hypotheses to repeatedly solve optimization problems in the form of factor graphs. 
Given suitable observations of one or multiple persons traversing the scene, the estimated camera poses converge towards a coherent extrinsic calibration within a few minutes.
We evaluate our approach in real-world settings and show that the calibration with our method achieves lower reprojection errors compared to a reference calibration generated by an offline method using a traditional calibration target.

\keywords{Camera calibration \hspace{-.5em} \and \hspace{-.5em} Human pose estimation \hspace{-.5em} \and \hspace{-.5em} Factor graphs}

\end{abstract}
\section{Introduction}
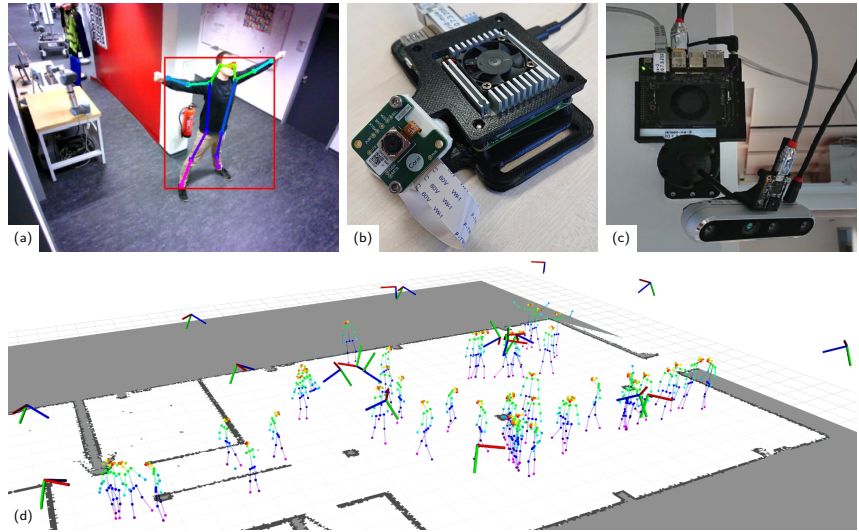
\begin{figure}[t]
		\centering
		\begin{tikzpicture}
			\node[inner sep=0,anchor=north west] (image1) at (0, 0) {\includegraphics[height=3.3cm]{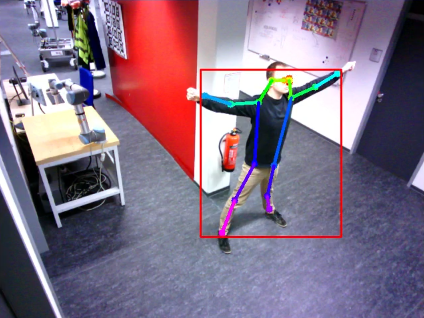}};
			\node[inner sep=0,anchor=north west,xshift=0.1cm] (image2) at (image1.north east) {\includegraphics[height=3.3cm]{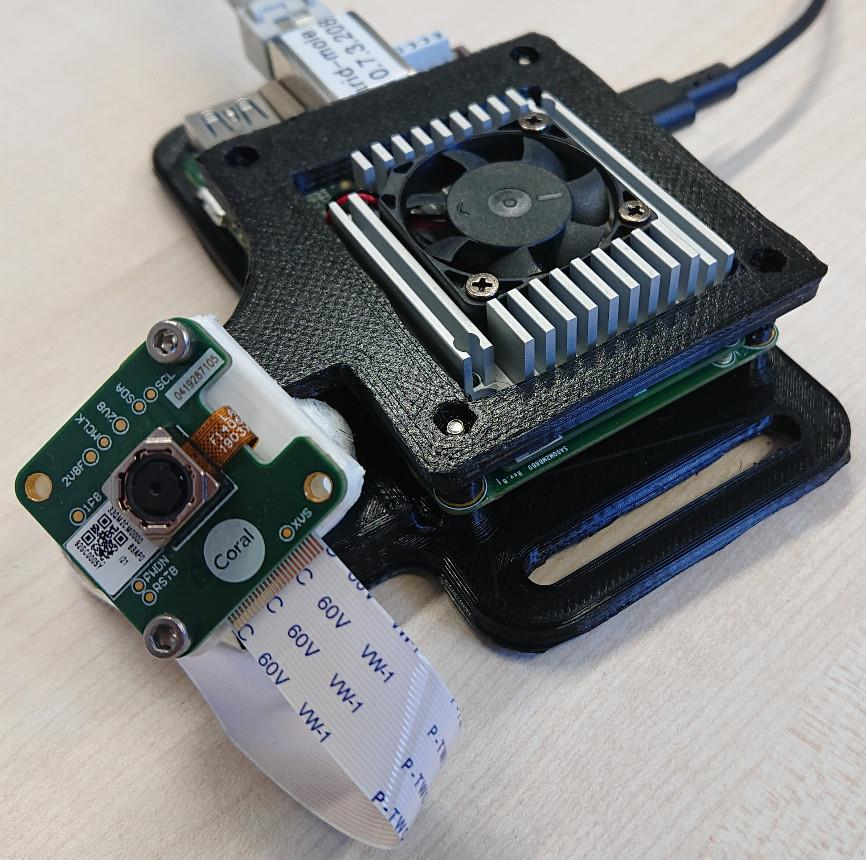}};
			\node[inner sep=0,anchor=north west,xshift=0.1cm] (image3) at (image2.north east) {\includegraphics[height=3.3cm]{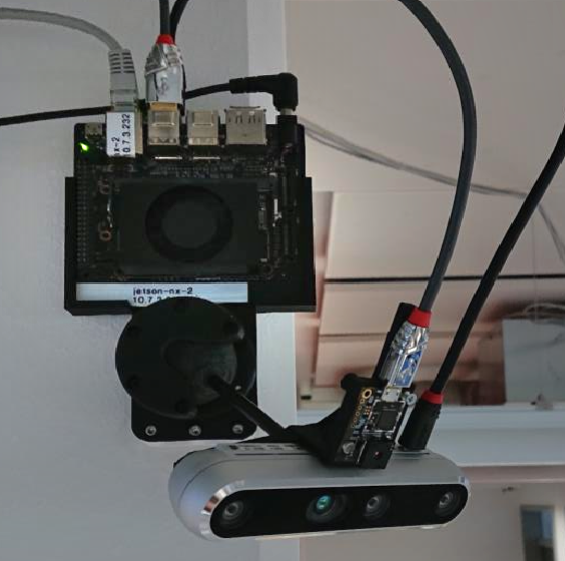}};
			\node[inner sep=0,anchor=north west,yshift=-0.1cm] (image4) at (image1.south west){\includegraphics[width=.925\textwidth]{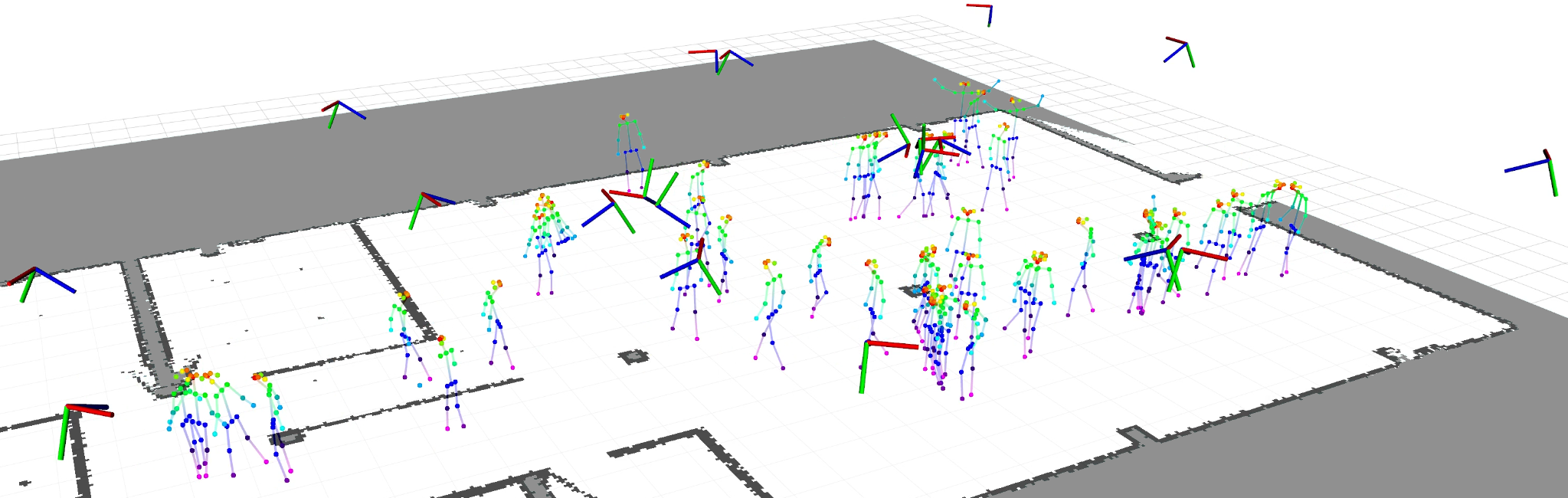}};
			
			\node[label,scale=.75, anchor=south west, rectangle, fill=white, align=center, font=\scriptsize\sffamily] (n_0) at (image1.south west) {(a)};
			\node[label,scale=.75, anchor=south west, rectangle, fill=white, align=center, font=\scriptsize\sffamily] (n_1) at (image2.south west) {(b)};
			\node[label,scale=.75, anchor=south west, rectangle, fill=white, align=center, font=\scriptsize\sffamily] (n_2) at (image3.south west) {(c)};
			\node[label,scale=.75, anchor=south west, rectangle, fill=white, align=center, font=\scriptsize\sffamily] (n_3) at (image4.south west) {(d)};
		\end{tikzpicture}			
		\vspace{-.7em}
		\caption{Extrinsic camera calibration using person keypoint detections (a) computed locally on different smart edge sensors: (b) based on a Google EdgeTPU Dev Board~\cite{edgetpu} and (c) based on an Nvidia Jetson Xavier NX Developer Kit~\cite{jetson}. 2D keypoints from multiple views are synchronized, filtered, and assigned to 3D person hypotheses (d) on a central backend. Observations are accumulated over time and a factor graph optimization is solved to obtain the optimal camera poses (coordinate systems in (d), with the blues axis being the view direction).}
		\label{fig:teaser}
		\vspace{-.7em}
\end{figure}

Sensor calibration is an essential prerequisite for most intelligent systems, as they combine data from numerous sensors to perceive a scene.
To successfully interpret and fuse the measurements of multiple sensors, they need to be transformed into a common coordinate frame.
This requires the sensor poses in a common reference frame---their extrinsic calibration.
An imprecise calibration can lead to degradation in performance and can possibly cause critical safety issues.  
	
For multiple reasons, the task of camera calibration is an inherently difficult one to solve~\cite{self_supervised}: 
First, the calibration parameters change over time, by normal usage, e.g. due to vibration, thermal expansion, or moving parts.
Therefore, it is not sufficient to calibrate the parameters only once during the construction of the system. Instead, calibration must be performed repeatedly throughout its lifetime.
Second, the calibration parameters cannot be measured directly with sufficient precision; they must be inferred from the data captured by the considered cameras.
Typically, the calibration is performed by actively deploying a calibration target of known correspondences in front of the cameras, e.g. a checkerboard pattern~\cite{zhang}.
However, this requires expertise and might be perceived as cumbersome and lengthy when it has to be applied repeatedly for a large multi-camera system.
Further challenges for inferring calibration parameters from image data involve accounting for noisy measurements and collecting a sufficient amount of data points spread over the entirety of the image planes.

In this work, we develop a novel, marker-free method for calibrating the extrinsic parameters of a network of static smart edge sensors, where each sensor runs inference for 2D human pose estimation.
In particular, we infer the relative poses between the cameras of the deployed smart edge sensor boards in real time using the person keypoint detections transmitted by the sensors~\cite{Simon}, as illustrated in \reffig{fig:teaser}. 2D keypoints from multiple views are synchronized, filtered, and assigned to 3D person hypotheses on a central backend, where observations are accumulated over time and a factor graph optimization~\cite{Dellaert} is solved to obtain the optimal camera poses.
The method can handle multiple persons in the scene, as well as arbitrary occlusions, e.g. from tables or pillars.
We assume the intrinsic parameters of the cameras to be known and a rough initial estimate of the extrinsic calibration to be available, which can easily be obtained, e.g. from a floor plan or by tape measure.

Our proposed method alleviates many of the issues mentioned above: No specific calibration target is required; it suffices for one or several persons to walk through the scene.
The method handles data association between multiple observed persons and their unknown dimensions. %
We propose an efficient online algorithm that optimizes the camera poses on-the-fly, giving direct feedback on success and when enough data has been captured.
The calibration procedure thus can be repeated easily to account for parameter change over time, without expert knowledge.
Furthermore, person keypoints can be detected from a significantly larger range of viewing angles (e.g. front, back, or side-view) than the pattern of a classical checkerboard calibration target, which is well detected only from a frontal view. This facilitates the collection of a sufficient amount of corresponding data points visible in multiple cameras that well constrain the factor graph optimization, further reducing the time required for calibration.

We evaluate the proposed approach in real-world settings and show that the calibration obtained with our method achieves better results in terms of reprojection errors in the targeted application domain of 3D multi-person pose estimation, compared to a reference calibration generated by an offline method using a traditional calibration target.
We make our implementation publicly available\footnote{\url{https://github.com/AIS-Bonn/ExtrCamCalib_PersonKeypoints}}.

\section{Related Work}
\paragraph*{\textbf{Camera Calibration.}}
Traditional methods for camera calibration are based on using artificial image features, so-called \textit{fiducials}. 
Their common idea is to deploy a calibration target with known correspondences in the overlapping field of view (FoV) of the considered cameras.
Zhang~\cite{zhang} utilizes a checkerboard pattern on a planar surface to perform intrinsic calibration of single cameras.
The \textit{kalibr} toolkit~\cite{kalibr_1} uses a planar grid of \textit{AprilTags}~\cite{AprilTag} to perform offline extrinsic and intrinsic calibration of multiple cameras, which allows to fully resolve the target's orientation towards the cameras and is robust against occlusions. 
We apply this method to obtain a reference calibration for evaluating our work.

Reinke \etal\cite{dog} propose an offline method for finding the relative poses between a set of (two) cameras and the base frame of a quadruped robot. They use a fiducial marker mounted on the end-effector of a limb as the calibration target. 
The camera poses are resolved using a \textit{factor graph}~\cite{Dellaert}, modeling kinematic constraints between the marker frame and the base frame together with the visual constraints.
We take up the idea of using factor graphs to model the calibration constraints in our work.

To cope with the issues of traditional approaches, methods for camera calibration have been proposed that do not extract fiducial features from calibration targets but use naturally occurring features instead.
Komorowski \etal\cite{stereo_view} extract SIFT features~\cite{SIFT} and find correspondences between multiple views using RANSAC~\cite{RANSAC_PnP}. They use segmentation to remove dynamic objects and validate their approach on stereo vision datasets. Their method is targeted towards one or a few small-baseline stereo cameras and offline processing of a small batch of images.
Bhardwaj \etal\cite{cars} calibrate traffic cameras by extracting vehicle instances via deep neural networks (DNNs) and matching them to a database of popular car models.
The extracted features and known dimensions of the car models are then used to formulate a P$n$P problem~\cite{RANSAC_PnP}. They assume a planar ground surface in the vicinity of the cars and process results offline.

A variety of methods considering surveillance scenarios use pedestrians as calibration targets.
Lv \etal\cite{lv} track head \& feet detections of a single pedestrian walking on a planar surface during the leg-crossing phases to perform offline extrinsic calibration of a single camera based on the geometric properties of vanishing points.
Following a tracking approach, they resolve the corresponding intrinsic parameters based on Zhang~\cite{zhang}.
Hödlmoser \etal\cite{hoedlmoser} use a similar approach as \cite{lv}, but expand the method to calibrate a camera network from pairwise relative calibrations.
The absolute scale of the camera network is resolved by manually specifying the height of the walking person.
Liu \etal\cite{liu1} require a moderately crowded scene to perform online intrinsic and extrinsic calibration of a single camera by assuming strong prior knowledge regarding the height distribution of the observed pedestrians.
The approach is based on computing vanishing points using RANSAC.
In \cite{liu2} they expand their method by introducing a joint calibration for a network of cameras based on the Direct Linear Transform~\cite{Zisserman}.
Henning \etal\cite{bodyslam} jointly optimize the trajectory of a monocular camera and a human body mesh by formulating an optimization problem in the form of a factor graph.
They apply a human motion model to constrain sequential body postures and to resolve scale.

Guan \etal\cite{pedestrians1,pedestrians2} detect head \& feet keypoints for each observable pedestrian and perform pairwise triangulation assuming an average height for all visible persons in the image pair.
They then compute the calibration offline, using RANSAC, followed by a gradient descent-based refinement scheme.
Their resulting calibration is only defined up to an unknown scale factor, which must be resolved manually. 
The method assumes the center lines between all pedestrians to be parallel, in other words, all persons are assumed to stand upright during the calibration, whereas other poses, e.g. sitting persons, are not supported.
Our method, in contrast, extracts up to 17 keypoints per person~\cite{COCO} using convolutional neural networks (CNNs) and assumes neither the dimensions or height of the persons, nor their pose or orientation towards the cameras to be known.
As for the unknown dimensions of the persons, the scale of our calibration is also ambiguous up to a single scale factor. To address this issue, we force the scale of the initial estimate of the extrinsic calibration to be maintained throughout the calibration procedure.

\paragraph*{\textbf{Human Pose Estimation.}}
Human pose estimation refers to the task of detecting anatomical keypoints of persons on images.
Early works use manually designed feature extractors like HOG-descriptors~\cite{hog_descriptors} or pictorial structures~\cite{pictorial_structures1,pictorial_structures2}.
In recent years, approaches using CNNs have become popular and yield impressive results.
Two well-known state-of-the-art and publicly available methods for human pose estimation are \textit{OpenPose}~\cite{openpose1} and \textit{AlphaPose}~\cite{alphapose2}.
Both methods estimate the poses of multiple persons simultaneously and in real time.

2D keypoint detections from multiple, calibrated camera views can be fused to obtain 3D human poses.
We consider a network of smart edge sensors, introduced by Bultmann \etal\cite{Simon,Simon2}, performing 3D human pose estimation in real time.
Each smart edge sensor performs 2D human pose estimation, processing the image data locally on the sensor boards and transmitting only the obtained keypoint data.
A central backend fuses the data received by the sensors to perform 3D human pose estimation via direct triangulation, and a semantic feedback loop is implemented to improve the 2D pose estimation on the sensor boards by incorporating global context information.

We adopt the smart edge sensor network~\cite{Simon} for our work, using the 2D person keypoints detected by the sensors as calibration features and aim to improve and facilitate the camera calibration required for this application scenario.

\section{Method}
Our method uses the image streams of a multi-camera system $\mathcal{S}_N$ with $N>1$ projective cameras $\mathcal{C}_i$, $i\in[0\dotss N-1]$, to extract and maximize knowledge about the relative poses between all cameras in real-time, i.e. finding the translation $\vec{t}_{ij}\in\mathbb{R}^3$ and rotation $\vec{R}_{ij}\in\mathcal{SO}(3)\subset\mathbb{R}^{3\times 3}$ between all camera pairs $ij$, where the pose of the optical center of $\mathcal{C}_i$ is defined by
\begin{align}
\vec{C}_i = \begin{pmatrix} \vec{R}_i & \vec{t}_i \\ \vec{0} & 1 \end{pmatrix} \in \mathcal{SE}\left(3\right)\,.
\end{align}
We call this the \textit{extrinsic calibration} of the multi-camera system $\mathcal{S}_N$.
Without loss of generality, we chose the first camera $\mathcal{C}_0$ to be the origin of the global reference frame and set $\vec{C}_0 = \vec{I}_{4\times 4}$.
In its local coordinate system, the view direction of each camera is the $z$-axis.

\begin{figure}[t]
	\centering
	\resizebox{1.0\linewidth}{!}{%
\begin{tikzpicture}[font=\sffamily,on grid,>={Stealth[inset=0pt,length=4pt,angle'=45]}]

\tikzset{every node/.append style={node distance=3.0cm}}
\tikzset{img_node/.append style={minimum size=1.5em,minimum height=3em,align=center,scale=0.65}}
\tikzset{sensor_node/.append style={minimum size=1.5em,minimum height=3em,minimum width={width("Estimation") +1.0em},draw,align=center,scale=0.65,fill=blue!15!white}}
\tikzset{content_node/.append style={minimum size=1.5em,minimum height=3em,minimum width={width("Synchronization") +1.0em},draw,align=center,scale=0.65,fill=blue!15!white}}
\tikzset{que_node/.append style={minimum size=1.5em,minimum height=3em,minimum width={width("Synchronization") +1.0em},draw,dotted,align=center,scale=0.65,fill=blue!11!white}}
\tikzset{label_node/.append style={scale=0.5, near start}}
\tikzset{junction/.append style={circle, fill=black, minimum size=3pt, draw}}

\definecolor{red}{rgb}     {0.9,0.0,0.0}
\definecolor{green}{rgb}   {0.0,0.5,0.0}
\definecolor{blue}{rgb}    {0.0,0.0,0.5}
\definecolor{grey}{rgb}    {0.5,0.5,0.5}

\draw[thick, rounded corners, grey!20!white,fill] (-0.9, -0.55) -- (3.48, -0.55) -- (3.48, 0.55) -- (-0.9, 0.55) -- cycle;
\draw[thick, rounded corners, grey!20!white,fill] (-0.9, -0.55 - 1.95) -- (3.48, -0.55 - 1.95) -- (3.48, 0.55 - 1.95) -- (-0.9, 0.55 - 1.95) -- cycle;
\draw[thick, rounded corners, grey!20!white,fill] (4.93, 0.15) -- (10.03, 0.15) -- (10.03, -3.55) -- (4.93, -3.55) -- cycle;

\node(camera1)[img_node] at (0, 0) {\includegraphics[width=1.6cm]{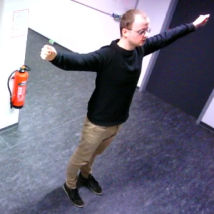}};
\node(posedet1)[sensor_node, right=2.5cm of camera1]{\large 2D Pose\\\large Estimation};
\node(skel1)[img_node, above right=+0.1cm and 1.8cm of posedet1] {\large 2D Pose\\[4pt]
	\setlength{\fboxsep}{0.5pt}%
	\setlength{\fboxrule}{0.5pt}%
	\fbox{\includegraphics[width=1.3cm]{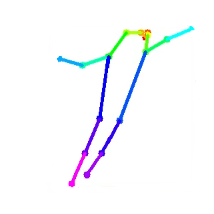}}};

\node(cameraN)[img_node, below of=camera1] {\includegraphics[width=1.6cm]{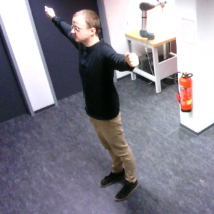}};
\node(posedetN)[sensor_node, right=2.5cm of cameraN]{\large 2D Pose\\\large Estimation};
\node(skelN)[img_node, above right=0.44cm and 1.8cm of posedetN] {\setlength{\fboxsep}{0.5pt}%
	\setlength{\fboxrule}{0.5pt}%
	\fbox{\includegraphics[width=1.3cm]{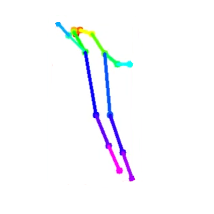}}};

\node(posedet_dummy) at ($(posedet1)!0.4!(posedetN)$){};

\node(sync)[content_node, above right=0.0cm and 3.73cm of posedet_dummy]{\large Synchronization};
\node(filter)[content_node, right=2.5cm of sync]{\large Preprocessing};

\node(que)[que_node, below=1.1cm of sync]{\large Queue};
\node(association)[content_node, right=2.5cm of que]{\large Data\\\large Association};

\node(opt)[content_node, below=1.1cm of que]{\large Optimization};
\node(refine)[content_node, right=2.5cm of opt]{\large Refinement};

\node(out)[above right=-0.4cm and 3.4cm of filter, scale=0.65, align=center]{\includegraphics[width=5.6cm]{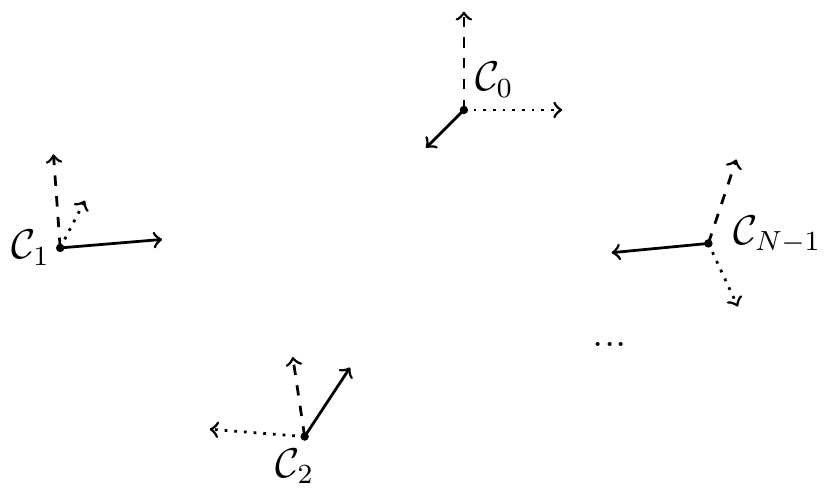}\\\large Extrinsic Calibration};

\draw[->, thick] (camera1) -- node[label_node,midway,above] {\large RGB} node[label_node,midway,below] {\large Image} (posedet1);
\draw[->, thick] (cameraN) -- node[label_node,midway,above] {\large RGB} node[label_node,midway,below] {\large Image} (posedetN);

\draw[->, thick] (posedet1) -| ++(1.18, -0.5) |- ([yshift= 2mm]sync.west);
\draw[->, thick] (posedetN) -| ++(1.18,  0.5) |- ([yshift=-2mm]sync.west);
\draw[->, dashed, thick] (posedet_dummy) + (1.0,0) -- (sync);
\draw[dotted, thick] ([yshift=-6.7mm]$(camera1)!0.5!(posedet1)$) -| ([yshift=8mm]$(cameraN)!0.5!(posedetN)$);

\draw[->, thick] (sync) -- (filter);
\draw[->, thick] (filter) -- (association);
\draw[->, thick] (association) -- (que);
\draw[->, thick] (que) -- (opt);
\draw[->, thick] (opt) -- (refine);
\draw[->, thick] (refine.east) -| (out.south);

\node[scale=0.6, anchor=north west] at (-0.9, 0.85) {\large\textbf{Smart Edge Sensor $0$}};
\node[scale=0.6, anchor=north west] at (-0.9, 0.85 - 1.95) {\large\textbf{Smart Edge Sensor $N-1$}};
\node[scale=0.6, anchor=north west] at (4.98, 0.1) {\large\textbf{Backend}};

\end{tikzpicture}
} %
	\vspace{-2em}
	\caption{Proposed pipeline for extrinsic camera calibration using smart edge sensors and person keypoint detections.
		Images are analyzed locally on the sensor boards.
		Keypoint detections are transmitted to the backend where multiple views are fused to construct and solve optimization problems using factor graphs.
		A queue decouples the preprocessing and optimization stages.}
	\vspace{-1em}
	\label{fig:pipeline}
\end{figure}
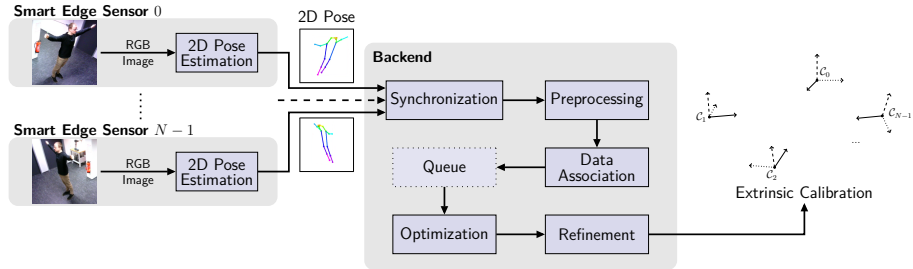
\reffig{fig:pipeline} gives an overview of our proposed pipeline. Each camera stream is fed into a person keypoint detector on the connected inference accelerator~\cite{Simon}. We refer to the unity of camera and detector as a \textit{smart edge sensor}.
The keypoint detections are transmitted to a central backend where they are time-synchronized and processed further.
The clocks of sensors and backend are software-synchronized via NTP and each keypoint detection message includes a timestamp representing the capture time of the corresponding image.

The \textit{preprocessing} stage removes redundant and noisy detections after which \textit{data association} is performed, where correspondences between person detections from multiple views are established.
Corresponding person detections are fused to form a person hypothesis and attached to a queue, which serves to decouple the preprocessing stage from the rest of the pipeline.
The \textit{optimization} stage continuously reads from this queue, selects several person hypotheses, and uses them to construct and solve an optimization problem in the form of a factor graph~\cite{Dellaert}.
The \textit{refinement} stage updates the current estimate of the extrinsic calibration by smoothing the intermediate results generated by the optimization and compensates for scaling drift w.r.t. the initialization.
As prerequisites for our method, we assume the intrinsic parameters of the cameras to be known and a rough initial estimate of the extrinsic calibration to be available, e.g. by tape measure or from a floor plan. %
The FoVs of all cameras must overlap in such a way that $\mathcal{S}_N$ forms a connected graph.

\subsection{Preprocessing}
The backend receives $N$ person keypoint detection streams and synchronizes and preprocesses them such that they can be used for optimization.
Each keypoint detection $\mathcal{D}_j^p$ is associated to a person instance $p$ and defined as
\begin{align} \label{eq:detection}
		\mathcal{D}_j^p=\{\,(u,v)^\mathsf{T},\,c\,,\,\boldsymbol{\Sigma}\,\}\,,
\end{align}
where $(u,v)^\mathsf{T}$ are the image coordinates, $c\in[0,1]$ is the confidence, $\boldsymbol{\Sigma} \in \mathbb{R}^{2\times2}$ is the covariance of the detection, and $j$ is the joint index. The covariance $\boldsymbol{\Sigma}$ is determined from the heatmaps used for keypoint estimation~\cite{Simon}.
First, incoming streams are synchronized into sets of time-corresponding detection messages of size $N$, which we will refer to as \textit{framesets} in the following.
Preprocessing then rejects false, noisy, redundant, and low-quality detections, passing through only detections that are considered accurate and suitable for contributing to improving the extrinsic calibration.
For this, we check different conditions for each frameset, which address the number of detections per sensor, the timestamps associated to each sensor, or the confidence value of each detection.
In particular, we reject all framesets where the maximum span or standard deviation of timestamps exceeds a threshold and consider only joint detections with a minimum confidence of $0.6$. We further require the hip and shoulder detections of each person to be valid, which is necessary for robust data association. %
After filtering, we use the distortion coefficients of each camera to undistort the coordinates of all valid detections using the \textit{OpenCV} library~\cite{opencv}.
   
\subsection{Data Association}
\begin{figure}[t]
	\centering
	\includegraphics[width=1.0\linewidth]{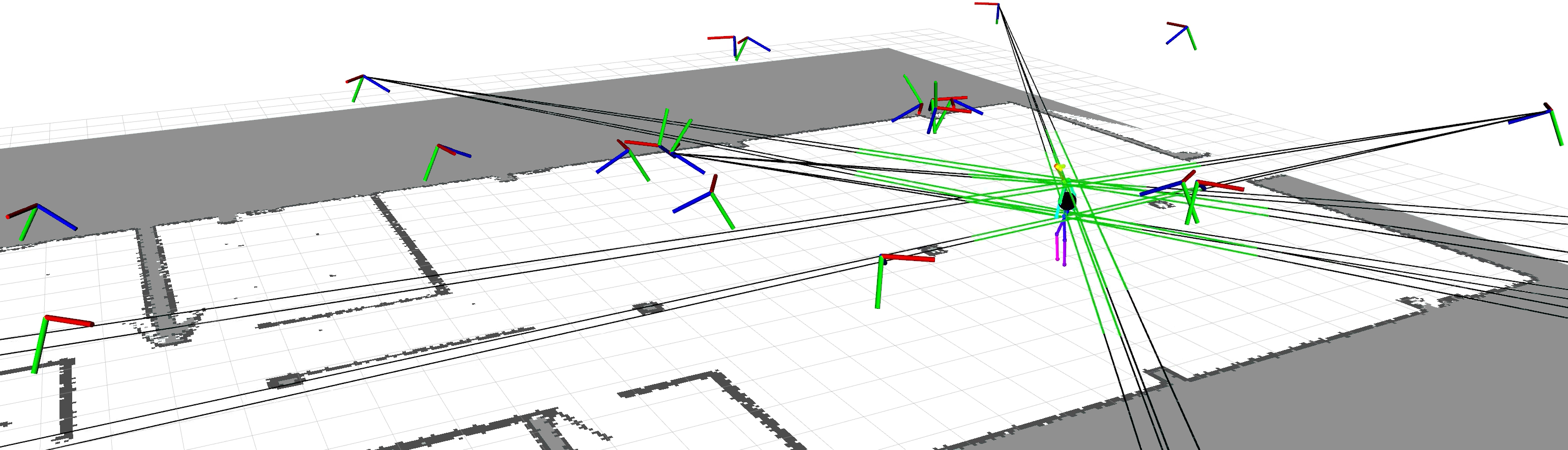}
	\vspace{-2em}
	\caption{Data Association: 3D back-projection rays embedded in the global coordinate system for the joint detections of one person (black), 
		the corresponding reduction to line segments after applying depth estimation (green), and the center of mass of the corresponding person hypothesis (black). 3D human pose estimation according to~\cite{Simon} shown for illustration purposes only.}
	\vspace{-1em}
	\label{fig:rays}
\end{figure}
In the data association step, we find correspondences between detections from different sensors based on the current estimate of the multi-view geometry of the camera network, which can still be inaccurate.
First, we back-project each 2D detection $\mathcal{D}$ into 3D, obtaining a ray $\vec{p}_\mathcal{D}$ with undetermined depth originating at the optical center of the respective camera.
Next, we reduce each ray $\vec{p}_\mathcal{D}$ to a line segment $\vec{l}_\mathcal{D}$ by estimating the interval $[z_\text{min},z_\text{max}]$ in which the depth $z$ of each detection $\mathcal{D}$ lies, as illustrated in \reffig{fig:rays}.
For the depth interval estimation, we assume a minimum and maximum torso height and width for the detected persons, derived by a specified minimum and maximum person height to be expected during calibration.
The four torso keypoints (shoulders and hips) empirically are the most stable and least occluded ones, and the physical distances between them can be assumed constant for a person due to the human anatomy, independent of the pose. 
The respective measured distance, however, depends on the persons' orientation towards the cameras. Here, we assume worst-case orientations, leading to larger depth intervals. 
In summary, we do not require persons to always stand upright but support arbitrary poses and orientations towards the cameras instead.
Specifying a short person height interval leads to a more constrained search space during data association, accommodating for an inaccurate initial estimate of the extrinsic calibration or a crowded scene.
A wider interval, however, yields equal results in common scenarios, while supporting small and tall persons as calibration targets alike. %
Depth estimation is the only component in the pipeline, where human anatomy is exploited.

To find the correspondences between person detections from multiple views, we deploy an iterative greedy search method similar to the approach of Tanke \etal\cite{greedy_matching}, using the distances between the previously estimated line segments as data association cost.
We define the distance of two line segments $\vec{l}_1$, $\vec{l}_2$ as the Closest Point-distance described by Wirtz \etal\cite{lines}:
\begin{align}
					d_{\text{closestpoint}}(\vec{l}_1,\vec{l}_2)=\min\left(d_\perp\left(\vec{l}_1,\vec{l}_2\right),d_\perp\left(\vec{l}_2,\vec{l}_1\right)\right)\,.
\end{align}

To further improve the robustness of the approach, as the extrinsic calibration is not precisely known, we iterate over all person detections, sorted in ascending depth order,
utilizing the depth estimation $(z_\text{min}+z_\text{max})/2$ of each person detection.
This exploits the fact that near person detections have a relatively short interval $[z_\text{min},z_\text{max}]$ and, thus, a more constrained localization in 3D space.
Lastly, we compute the \textit{center~of~mass} for each person hypothesis, which serves to give a rough localization of each person hypothesis in 3D space, by averaging the center points of the line segments of all assigned torso keypoints.

The data association stage outputs a list of person hypotheses that have been observed by at least two different cameras, which will be used to construct a factor graph optimization problem in the following.

\subsection{Factor Graph Optimization}
The optimization stage processes the person hypotheses obtained through data association to extract knowledge about the extrinsic calibration of the utilized multi-camera system.
To this end, we construct a factor graph~\cite{GTSAM} encoding projective constraints, based on a selection of person hypotheses, as well as prior knowledge on the camera poses w.r.t. the initial estimate of the extrinsic calibration or the results of previous optimization cycles.

The optimal selection of person hypotheses used in an optimization cycle is determined by a selection algorithm from all available hypotheses, ensuring an optimal spatial and temporal distribution of the observations, to obtain a well-constrained optimization problem while also maintaining a reasonable degree of entropy between selections over consecutive optimization cycles.
For this, we generate a random permutation of the indices of all available person hypotheses, which is biased towards selecting newer hypotheses first.
Selecting newer hypotheses with higher probability is advantageous, as their data association and center of mass are estimated more reliably, given that the extrinsic calibration improves over time.
Additionally, we ensure a minimum spacing between all selected person hypotheses, w.r.t. to their center of mass, by only including the next person hypothesis within the permutation if its distance towards all previously selected person hypotheses is above a spacing threshold $s=\SI{0.2}{\meter}$.

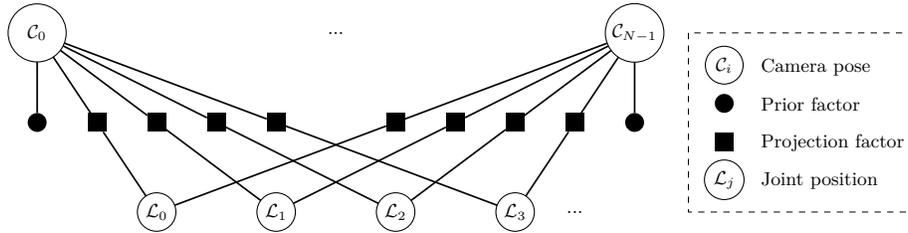
\begin{figure}[t]
	\centering
	\resizebox{1.\linewidth}{!}{%
\begin{tikzpicture}
	
	\Vertex[L=$\mathcal{L}_0$,x=0,y=0]{l0}
	\Vertex[L=$\mathcal{L}_1$,x=2,y=0]{l1}
	\Vertex[L=$\mathcal{L}_2$,x=4,y=0]{l2}
	\Vertex[L=$\mathcal{L}_3$,x=6,y=0]{l3}

	\Vertex[L=$\mathcal{L}_j$,x=9.5,y=0.54]{l99}

	\Vertex[L=${~}~\mathcal{C}_{0}~{~}$,x=-2,y=+3]{c0}
	\Vertex[L=$\mathcal{C}_{N-1}$,x=8,y=+3]{c2}
	
	\Vertex[L=$\mathcal{C}_{i}$,x=9.5,y=+2.46]{c99}

	\tikzset{VertexStyle/.style = {shape = rectangle,fill = black, minimum size=9pt}}
	
	\Vertex[L=~,x=-1,y=+1.5,LabelOut=true,Lpos=90,Ldist=1.1]{f0}
	\Vertex[L=~,x=+0,y=+1.5,LabelOut=true,Lpos=90,Ldist=1.1]{f1}
	\Vertex[L=~,x=+1,y=+1.5,LabelOut=true,Lpos=90,Ldist=1.1]{f2}
	\Vertex[L=~,x=+2,y=+1.5,LabelOut=true,Lpos=90,Ldist=1.1]{f3}

	\Vertex[L=~,x=+4,y=+1.5,LabelOut=true,Lpos=90,Ldist=1.1]{f9}
	\Vertex[L=~,x=+5,y=+1.5,LabelOut=true,Lpos=90,Ldist=1.1]{f10}
	\Vertex[L=~,x=+6,y=+1.5,LabelOut=true,Lpos=90,Ldist=1.1]{f11}
	\Vertex[L=~,x=+7,y=+1.5,LabelOut=true,Lpos=90,Ldist=1.1]{f12}

	\Edge(c0)(f0)
	\Edge(c0)(f1)
	\Edge(c0)(f2)
	\Edge(c0)(f3)
	\Edge(f0)(l0)
	\Edge(f1)(l1)
	\Edge(f2)(l2)
	\Edge(f3)(l3)

	\Edge(c2)(f9)
	\Edge(c2)(f10)
	\Edge(c2)(f11)
	\Edge(c2)(f12)
	\Edge(f9)(l0)
	\Edge(f10)(l1)
	\Edge(f11)(l2)
	\Edge(f12)(l3)
	
	\Vertex[L=~,x=+9.5,y=+1.18,LabelOut=true,Lpos=0,Ldist=1.1]{f8}

	\tikzset{VertexStyle/.style = {shape = circle,fill = black, minimum size=9pt}}
	\Vertex[L=~,x=-2.0,y=+1.5,LabelOut=true,Lpos=90,Ldist=1.1]{p0}
	\Vertex[L=~,x=+8.0,y=+1.5,LabelOut=true,Lpos=90,Ldist=1.1]{p3}
	\Edge(c0)(p0)
	\Edge(c2)(p3)

	\Vertex[L=~,x=+9.5,y=+1.82,LabelOut=true,Lpos=0,Ldist=1.1]{p2}
	\draw[draw=black,dashed] (8.9,3) rectangle ++(3.85,-3);

	\node[anchor=west] (NC) at (10.0,0.54-0.01) {Joint position};
	\node[anchor=west] (NC) at (10.0,2.46-0.04) {Camera pose};

	\node[anchor=west] (NC) at (10.0,1.82-0.02) {Prior factor};
	\node[anchor=west] (NC) at (10.0,1.18-0.02) {Projection factor};
	
	\node (NC) at (3,+3) {\footnotesize ...};
	\node (NL) at (7,0) {\footnotesize ...};

\end{tikzpicture}
} %
	\vspace{-2em}
	\caption{Factor graph with camera variable nodes for the camera poses~$\mathcal{C}_i$ and landmark variable nodes for the 3D person joint positions~$\mathcal{L}_j$. 
		Camera and landmark nodes can be connected via binary projection factors to constrain the reprojection error of a person keypoint detection.
		Each landmark node must be connected to at least two projection factors for allowing triangulation.
		All camera nodes are connected to a unary prior factor that encodes the initial uncertainty of the camera pose.}
	\vspace{-1em}
	\label{fig:graph_method}
\end{figure}

For each optimization cycle $t$, we construct a factor graph $\mathcal{G}_t$ by using a selection of person hypotheses $\mathcal{H}_t$.
A factor graph is a bipartite graph consisting of \textit{variable nodes} and \textit{factor nodes}~\cite{Dellaert}.
Variable nodes represent the unknown random variables of the optimization problem, i.e. the 3D joint positions of the person hypotheses in $\mathcal{H}_t$ (\textit{landmark nodes}) and the considered camera poses of the multi-camera system $\mathcal{S}_N$ (\textit{camera nodes}).
Factor nodes constrain the variable nodes by encoding the available knowledge about the underlying distribution of the considered random variables.
In particular, this refers to the obtained observations contained in $\mathcal{H}_t$ as well as the resulting camera poses from previous optimization cycles. 
Each factor node uses a Gaussian noise model that reflects the confidence in the constraint it represents.
The constructed factor graph is illustrated in \reffig{fig:graph_method}.
We equip every camera node $\mathcal{C}_i^t$ with a unary prior factor, encoding prior knowledge about the camera pose and its uncertainty, and use binary projection factors connecting camera nodes to landmark nodes to encode observation constraints based on person keypoint detections.
Projection factors calculate the reprojection error for a 2D detection w.r.t. the corresponding camera pose and landmark position using the known intrinsic parameters $\vec{K}_i$.
Camera nodes are initialized with the current estimate for the extrinsic calibration (for $t=0$, we use the initial estimate and for $t>0$, we resuse the result of the previous time step).
Landmark nodes are initialized by triangulation of the 2D observations using the latest camera geometry estimate~\cite{Zisserman}.
Note, that we perform triangulation in every optimization cycle, even when using a person hypothesis that was already utilized in a previous optimization cycle.
Hence, the triangulation results are updated based on the current estimate of the extrinsic calibration.
We solve each factor graph $\mathcal{G}_t$ for the most likely camera poses by applying a Levenberg-Marquardt optimization scheme, provided by the GTSAM framework~\cite{GTSAM}.
A successful optimization yields a new candidate for the extrinsic calibration of $\mathcal{S}_N$. 
We forward this candidate to the refinement stage where the current estimate of the extrinsic calibration will be updated based on this candidate.
The updated estimate for the extrinsic calibration will then be used for constructing and initializing the factor graph in the next optimization cycle.

\subsection{Camera Pose Refinement}
After each successful optimization, we obtain new candidates $\hat{\mathcal{C}}_i^t$ for the extrinsic calibration of a subset of cameras $i\subseteq\mathcal{S}_N$ that were constrained by the factor graph $\mathcal{G}_t$.
We smooth between the previous state and the new measurement using a Kalman filter to obtain the current estimate of the extrinsic calibration $\mathcal{C}_i^t$. 
As each optimization cycle contains only a limited number of observations in the factor graph to enable real-time operation, smoothing prevents overconfidence towards a specific set of observations and improves the convergence behavior.
We update the previous estimate with the result of the factor graph optimization using the marginalized uncertainty from the optimized factor graph as measurement noise. 
Between optimization cycles, we add a constant process noise to each predicted camera pose uncertainty, enabling convergence over longer time horizons.
Finally, we prevent scaling drift of the updated extrinsic calibration by applying the scaling factor that minimizes the distance towards the initial estimate of the extrinsic calibration according to Umeyama's method~\cite{Umeyama}.

\section{Evaluation}
\begin{figure}[b]
	\vspace{-1.5em}
	\centering
	\begin{tikzpicture}
		\node[inner sep=0,anchor=north west] (image1) at (0, 0) {\includegraphics[width=0.75\linewidth]{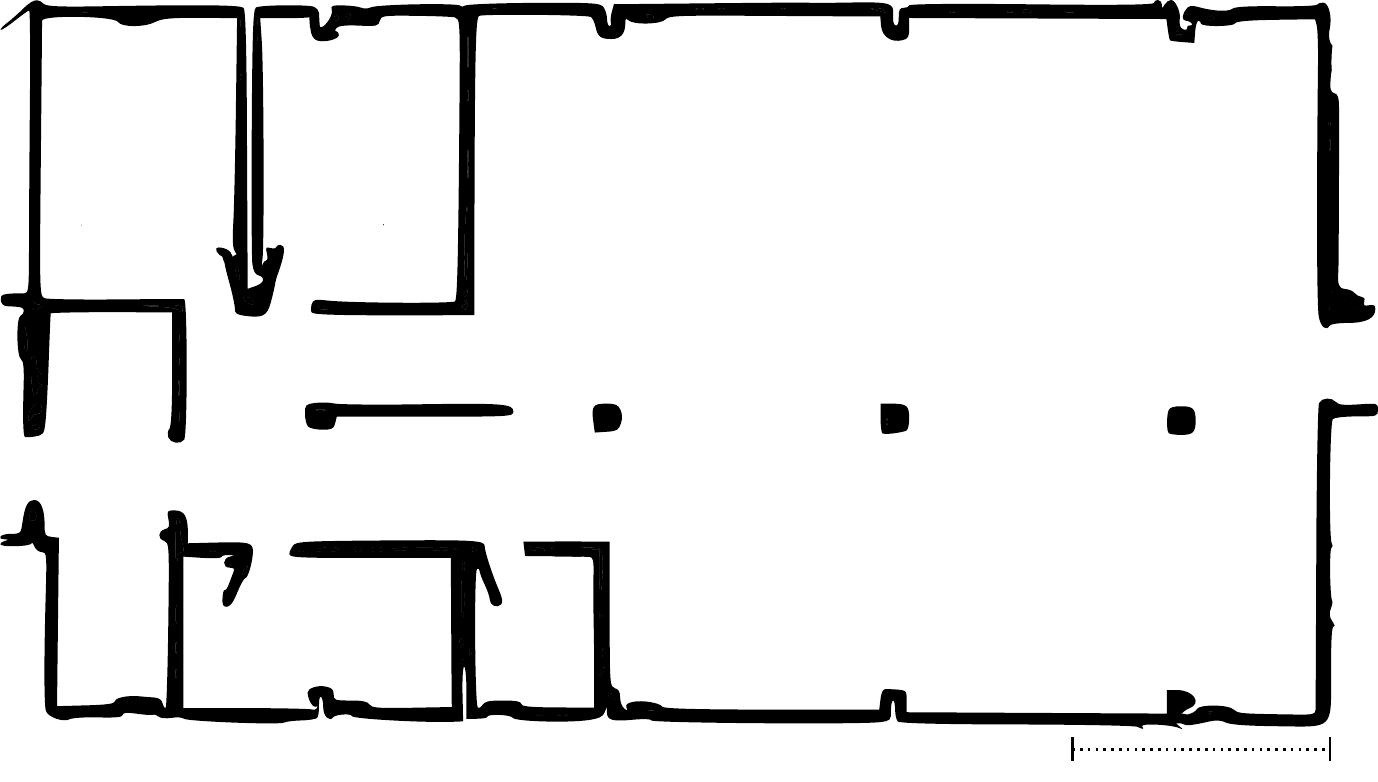}};
		\node[inner sep=0,anchor=north west] (image2) at (0, 0) {\includegraphics[page=2,width=0.75\linewidth]{mapwith20cams.pdf}};
		\begin{scope}[shift=(image1.south west),x={(image1.south east)},y={(image1.north west)}]
			\node[label,scale=1.0, anchor=north west, rectangle, align=center, font=\scriptsize\sffamily] (n_0) at (0.8,0.05) {5 meters};
			\node[label,scale=1.25, anchor=south west, rectangle, align=center, font=\scriptsize\sffamily] (c_0) at (0.6,0.5) {$\mathcal{C}_{0}$};
			\node[label,scale=1.25, anchor=south west, rectangle, align=center, font=\scriptsize\sffamily] (c_1) at (0.6,0.31) {$\mathcal{C}_{1}$};
			\node[label,scale=1.25, anchor=south west, rectangle, align=center, font=\scriptsize\sffamily] (c_2) at (0.43,0.33) {$\mathcal{C}_{2}$};
			\node[label,scale=1.25, anchor=south west, rectangle, align=center, font=\scriptsize\sffamily] (c_3) at (0.415,0.5) {$\mathcal{C}_{3}$};
			\node[label,scale=1.25, anchor=south west, rectangle, align=center, font=\scriptsize\sffamily] (c_4) at (0.33,0.3) {$\mathcal{C}_{4}$};
			\node[label,scale=1.25, anchor=south west, rectangle, align=center, font=\scriptsize\sffamily] (c_5) at (0.1,0.33) {$\mathcal{C}_{5}$};
			\node[label,scale=1.25, anchor=south west, rectangle, align=center, font=\scriptsize\sffamily] (c_6) at (0.16,0.51) {$\mathcal{C}_{6}$};
			\node[label,scale=1.25, anchor=south west, rectangle, align=center, font=\scriptsize\sffamily] (c_7) at (0.38,0.38) {$\mathcal{C}_{7}$};
			\node[label,scale=1.25, anchor=south west, rectangle, align=center, font=\scriptsize\sffamily] (c_8) at (0.34,0.62) {$\mathcal{C}_{8}$};
			\node[label,scale=1.25, anchor=south west, rectangle, align=center, font=\scriptsize\sffamily] (c_9) at (0.37,0.89) {$\mathcal{C}_{9}$};
			\node[label,scale=1.25, anchor=south west, rectangle, align=center, font=\scriptsize\sffamily] (c_10) at (0.56,0.89) {$\mathcal{C}_{10}$};
			\node[label,scale=1.25, anchor=south west, rectangle, align=center, font=\scriptsize\sffamily] (c_11) at (0.665,0.89) {$\mathcal{C}_{11}$};
			\node[label,scale=1.25, anchor=south west, rectangle, align=center, font=\scriptsize\sffamily] (c_12) at (0.67,0.5) {$\mathcal{C}_{12}$};
			\node[label,scale=1.25, anchor=south west, rectangle, align=center, font=\scriptsize\sffamily] (c_13) at (0.885,0.83) {$\mathcal{C}_{13}$};
			\node[label,scale=1.25, anchor=south west, rectangle, align=center, font=\scriptsize\sffamily] (c_14) at (0.88,0.63) {$\mathcal{C}_{14}$};
			\node[label,scale=1.25, anchor=south west, rectangle, align=center, font=\scriptsize\sffamily] (c_15) at (0.88,0.13) {$\mathcal{C}_{15}$};
			\node[label,scale=1.25, anchor=south west, rectangle, align=center, font=\scriptsize\sffamily] (c_16) at (0.69,0.1) {$\mathcal{C}_{16}$};
			\node[label,scale=1.25, anchor=south west, rectangle, align=center, font=\scriptsize\sffamily] (c_17) at (0.67,0.31) {$\mathcal{C}_{17}$};
			\node[label,scale=1.25, anchor=south west, rectangle, align=center, font=\scriptsize\sffamily] (c_18) at (0.56,0.1) {$\mathcal{C}_{18}$};
			\node[label,scale=1.25, anchor=south west, rectangle, align=center, font=\scriptsize\sffamily] (c_19) at (0.48,0.1) {$\mathcal{C}_{19}$};
		\end{scope}
	\end{tikzpicture}
	\vspace{-1em}
	\caption{Sketched floor plan with camera poses for the evaluation experiments.}
	\label{fig:maps}
\end{figure}

We evaluate the proposed method in challenging real-world settings in our lab, a large room with an area of $\sim$\SI{240}{\square\meter} and a height of \SI{3.2}{\meter}.
As the room is partly a robotics workshop and partly a desk-based workspace, it is densely filled with different objects and furniture, which can cause false detections and occlusion.
The cameras are distributed throughout the room as illustrated in \reffig{fig:maps} at similar heights of around \SI{2.6}{\meter}.
We deploy $20$ smart edge sensors, $16$ of which are based on the Google EdgeTPU, and $4$ are based on the Nvidia Jetson Xavier NX (cf. \reffig{fig:teaser}).
Both sensor types provide person keypoint detections in identical format and will be treated in the same way during the experiments.

For evaluation, we apply our pipeline to recordings of one and two persons (\SI{1.96}{\meter} \& \SI{1.70}{\meter} ) crossing the room and generating detections in all cameras over $\sim$\SI{180}{\second}. %
We repeat the experiment $10$ times with different initializations and compare our results towards a reference calibration obtained by applying the kalibr toolkit~\cite{kalibr_1}.
We apply an initial error of \SI{0.25}{\meter} and \SI{10}{\degree} in a random direction w.r.t. the reference calibration for all cameras $\mathcal{C}_i$ for $i>0$ and use default parameters provided in the linked repository.
We empirically verified that errors of this order of magnitude are easily attainable via manual initialization utilizing a floor plan, height measurements, and RGB images from all cameras.

\begin{table}[t]
	\centering
	\caption{Statistics of the position and orientation error towards the reference calibration averaged over $10$ repetitions of the experiment.}
	\vspace{-.5em}
	\setlength{\tabcolsep}{2pt}
	\begin{tabular}{ |c|S|S|S|S||S|S|S|S| }
		\cline{2-9}
		\multicolumn{1}{c|}{} & \multicolumn{4}{c||}{\textbf{1 Person}} & \multicolumn{4}{c|}{\textbf{2 Persons}}\\[0.02cm]
		\hline
		{\textbf{Error}}&
		{\textbf{Avg.}}&            
		{\textbf{Std.}}&
		{\textbf{Min.}}&
		{\textbf{Max.}}&
		{\textbf{Avg.}}&            
		{\textbf{Std.}}&
		{\textbf{Min.}}&
		{\textbf{Max.}}\\[0.05cm]
		\hline\hline
		{Position}&
		{\SI{0.053}{\meter}}&
		{\SI{0.030}{\meter}}&
		{\SI{0.011}{\meter}}&
		{\SI{0.119}{\meter}}&
		{\SI{0.052}{\meter}}&
		{\SI{0.030}{\meter}}&
		{\SI{0.017}{\meter}}&
		{\SI{0.122}{\meter}}\\      
		\hline
		{Orientation}& 
		{\SI{0.390}{\degree}}&    
		{\SI{0.184}{\degree}}&
		{\SI{0.120}{\degree}}&
		{\SI{0.891}{\degree}}&
		{\SI{0.436}{\degree}}&    
		{\SI{0.177}{\degree}}&
		{\SI{0.154}{\degree}}&
		{\SI{0.818}{\degree}}\\			
		\cline{0-8}      	
	\end{tabular}
	\vspace{-.3em}
	\label{tab:means}
\end{table}
\begin{figure}[t] %
	\centering
	\begin{tikzpicture}
		\node[inner sep=0,anchor=north west] (image1) at (0, 0) {\includegraphics[width=0.491\textwidth]{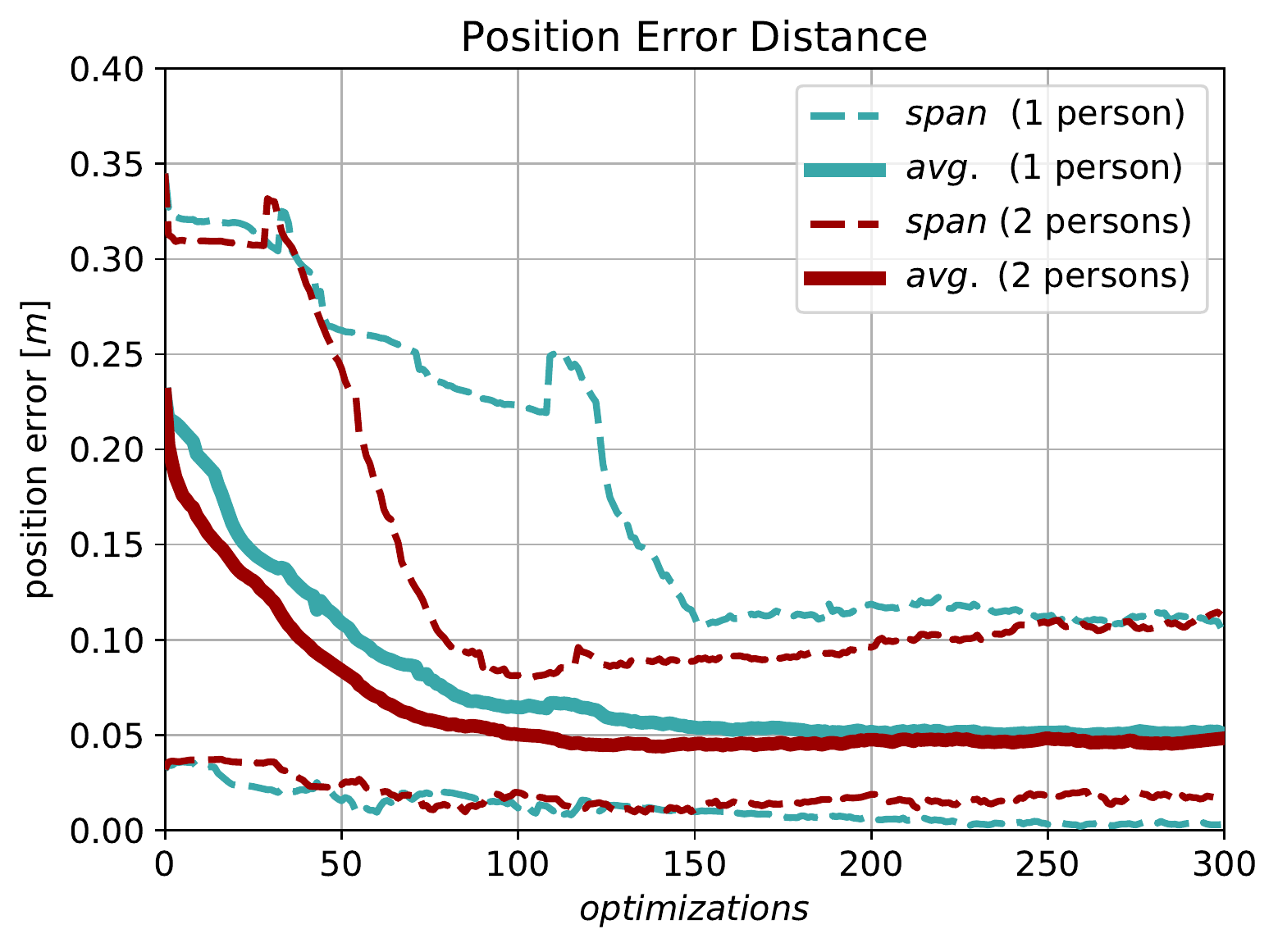}};
		\node[inner sep=0,anchor=north west,xshift=0.1cm] (image2) at (image1.north east) {\includegraphics[width=0.491\textwidth]{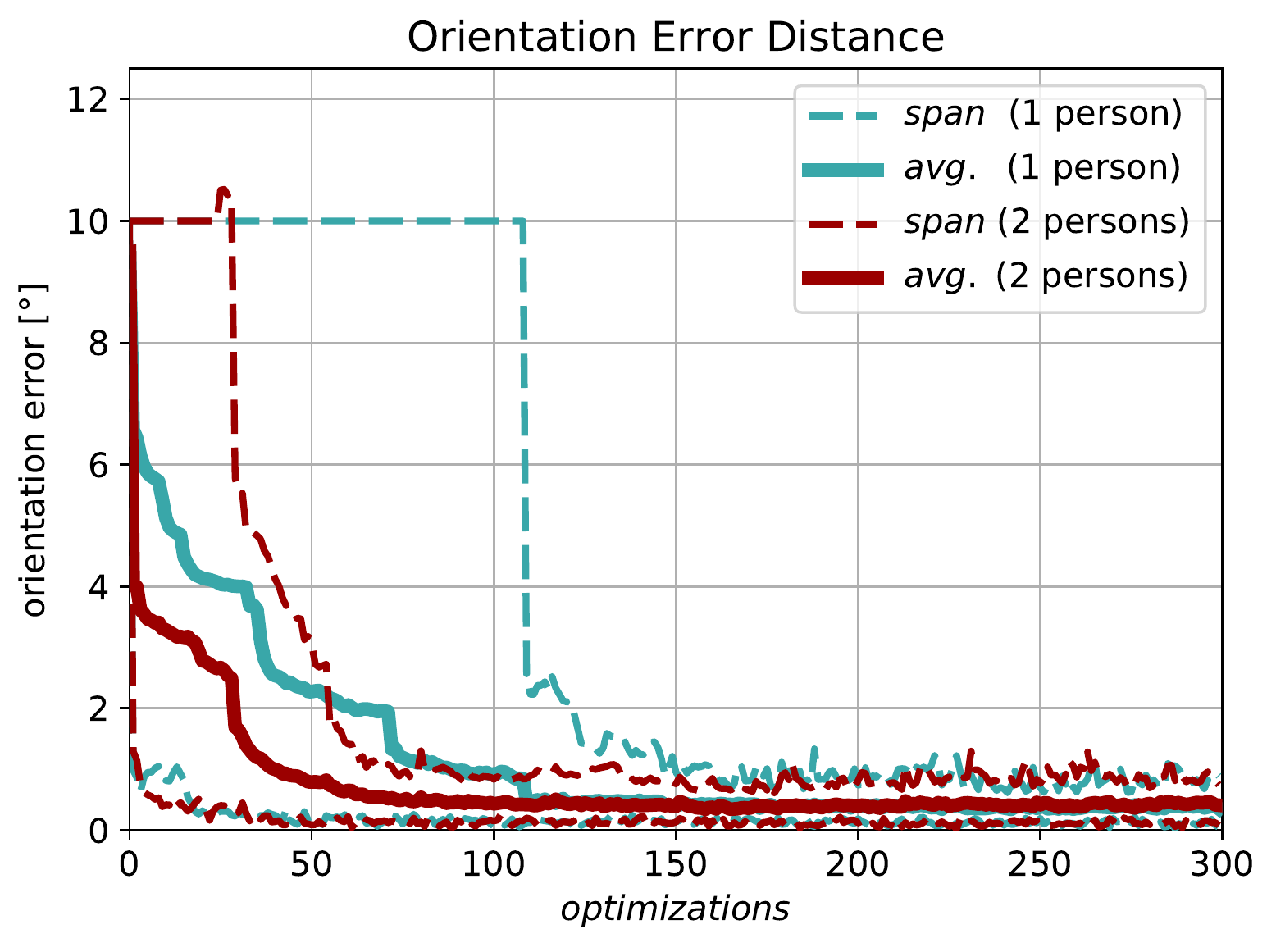}};

		\node[label,scale=1, anchor=south west, rectangle, fill=white, align=center, font=\scriptsize\sffamily] (n_0) at (image1.south west) {(a)};
		\node[label,scale=1, anchor=south west, rectangle, fill=white, align=center, font=\scriptsize\sffamily] (n_1) at (image2.south west) {(b)};
	\end{tikzpicture}				
	\vspace{-1.em}
	\caption{Evolution of mean and min--max span of (a) position and (b) orientation error towards the reference calibration. Convergence is faster when observing multiple persons.} %
	\label{fig:experiment4}
	\vspace{-1.em}
\end{figure}

\reftab{tab:means} shows the statistics of the final position and orientation error distributions towards the reference calibration averaged over all repetitions of the experiments with one or two persons present in the scene, respectively.
The position error is obtained by rigid alignment of the calibration result towards the reference according to Umeyama's method~\cite{Umeyama} without rescaling.
The orientation error is computed as the angle between two orientations via the shortest arc~\cite{quaternion_distance}.
We do not observe a significant difference in the final result between calibrating with one or two persons.
However, convergence is faster in the two-person case, as all cameras provide detections earlier in the procedure.

\reffig{fig:experiment4} shows the evolution of the error over time for one exemplary repetition of the resp. experiment with one or two persons.
The majority of the convergence takes place in the first ${\sim}50$ optimization cycles or $\sim$\SI{35}{\second} and after ${\sim}100$ optimizations, the camera poses and errors remain stable in the two-person experiment. With only a single person, convergence is slower. Observations from all cameras are obtained after ${\sim}110$ optimization cycles and it takes ${\sim}150$ iterations for the poses to remain stable.

\reftab{tab:init_errors} shows the final position error for different initialization errors with two persons.
Convergence remains stable for initial errors up to \SI{35}{\centi\meter} and \SI{15}{\degree} but becomes less reliable for larger errors.
In particular, the likelihood of the camera poses being stuck in a local minimum consistent with queued person hypotheses containing false data association increases with larger initialization errors, %
as the accuracy of the data association relies on the geometry of the provided initialization.
\begin{table}[t]
	\centering
	\caption{Position error towards the reference calibration for different initial errors.} %
	\setlength{\tabcolsep}{3pt}
	\begin{tabular}{ |c|c|c|c|c|c| }
		\hline
		{\textbf{Initial Error}}&
		{\textbf{\SI{0.10}{\meter}}, \textbf{\SI{5}{\degree}}}&
		{\textbf{\SI{0.25}{\meter}}, \textbf{\SI{10}{\degree}}}&
		{\textbf{\SI{0.35}{\meter}}, \textbf{\SI{15}{\degree}}}&
		{\textbf{\SI{0.50}{\meter}}, \textbf{\SI{20}{\degree}}}&
		{\textbf{\SI{0.75}{\meter}}, \textbf{\SI{20}{\degree}}}\\[0.05cm]
		\hline
		\hline
		{Final Error}&
		{\SI{0.050}{\meter}}&
		{\SI{0.052}{\meter}}&
		{\SI{0.067}{\meter}}&
		{\SI{0.118}{\meter}}&
		{\SI{0.214}{\meter}}\\
		\hline
		{Std.}&
		{\SI{0.003}{\meter}}&
		{\SI{0.030}{\meter}}&
		{\SI{0.011}{\meter}}&
		{\SI{0.073}{\meter}}&
		{\SI{0.164}{\meter}}\\	
		\cline{0-5}      	
	\end{tabular}
	\label{tab:init_errors}
\end{table}

Additionally, we compare reprojection errors, measured using two different evaluation pipelines, using the calibration obtained from our experiments with two persons.
The first evaluation processes keypoint detections for 3D human pose estimation~\cite{Simon} and matches the domain in which our calibration was obtained. Here, we use a distinct recording unseen during the calibration. %
The second pipeline uses a sequence of multi-view images of the AprilTag grid used to obtain the reference calibration~\cite{kalibr_1} and, thus, matches its data domain.
In general, the keypoint-based evaluation is biased towards our keypoint-based calibration method, while the AprilTag evaluation is biased towards the reference calibration.

\begin{figure}[t] %
	\centering
	\begin{tikzpicture}
		\node[inner sep=0,anchor=north west] (image1) at (0, 0) {\includegraphics[width=0.491\textwidth]{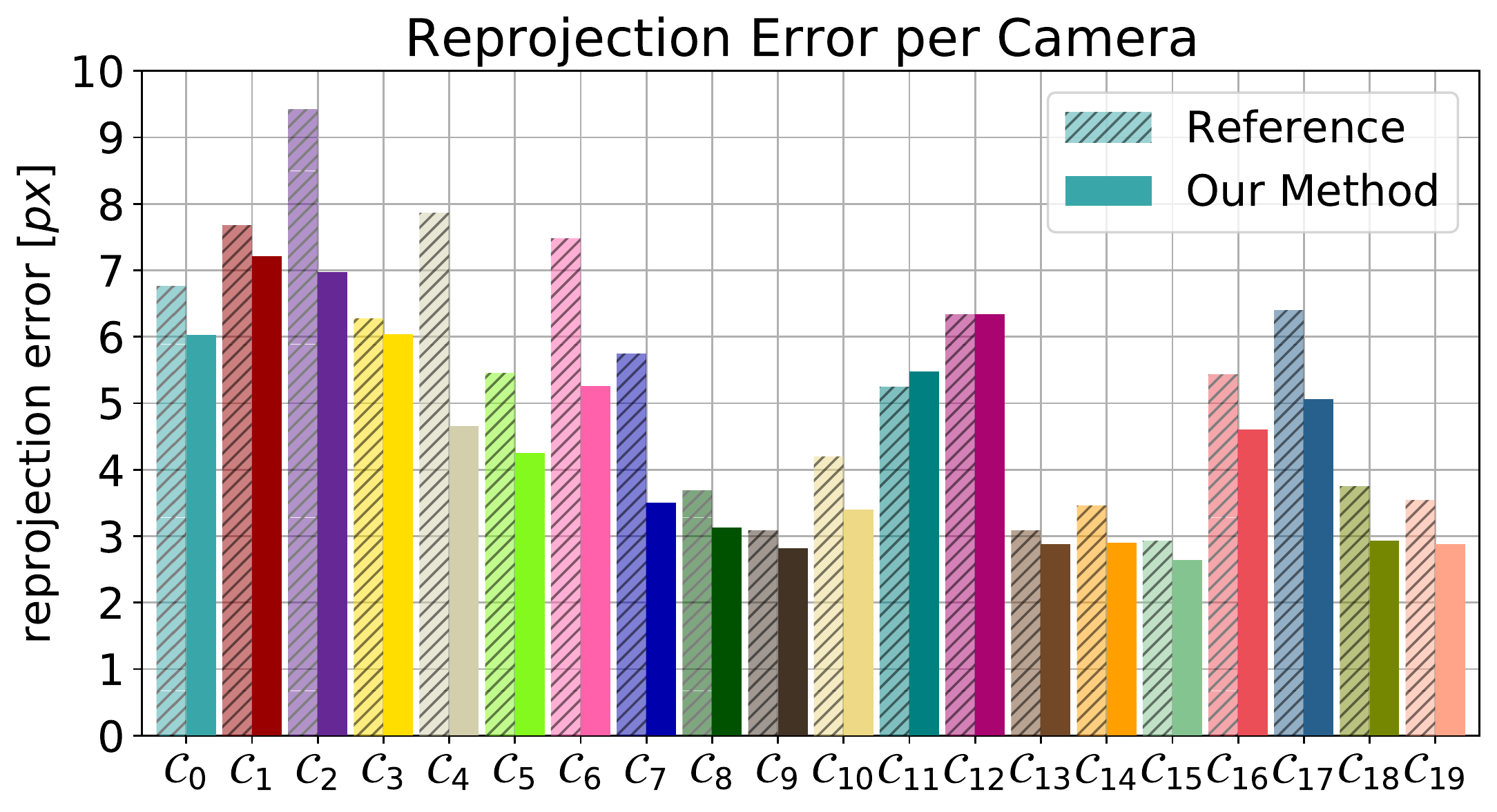}};
		\node[inner sep=0,anchor=north west,xshift=0.1cm] (image2) at (image1.north east) {\includegraphics[width=0.491\textwidth]{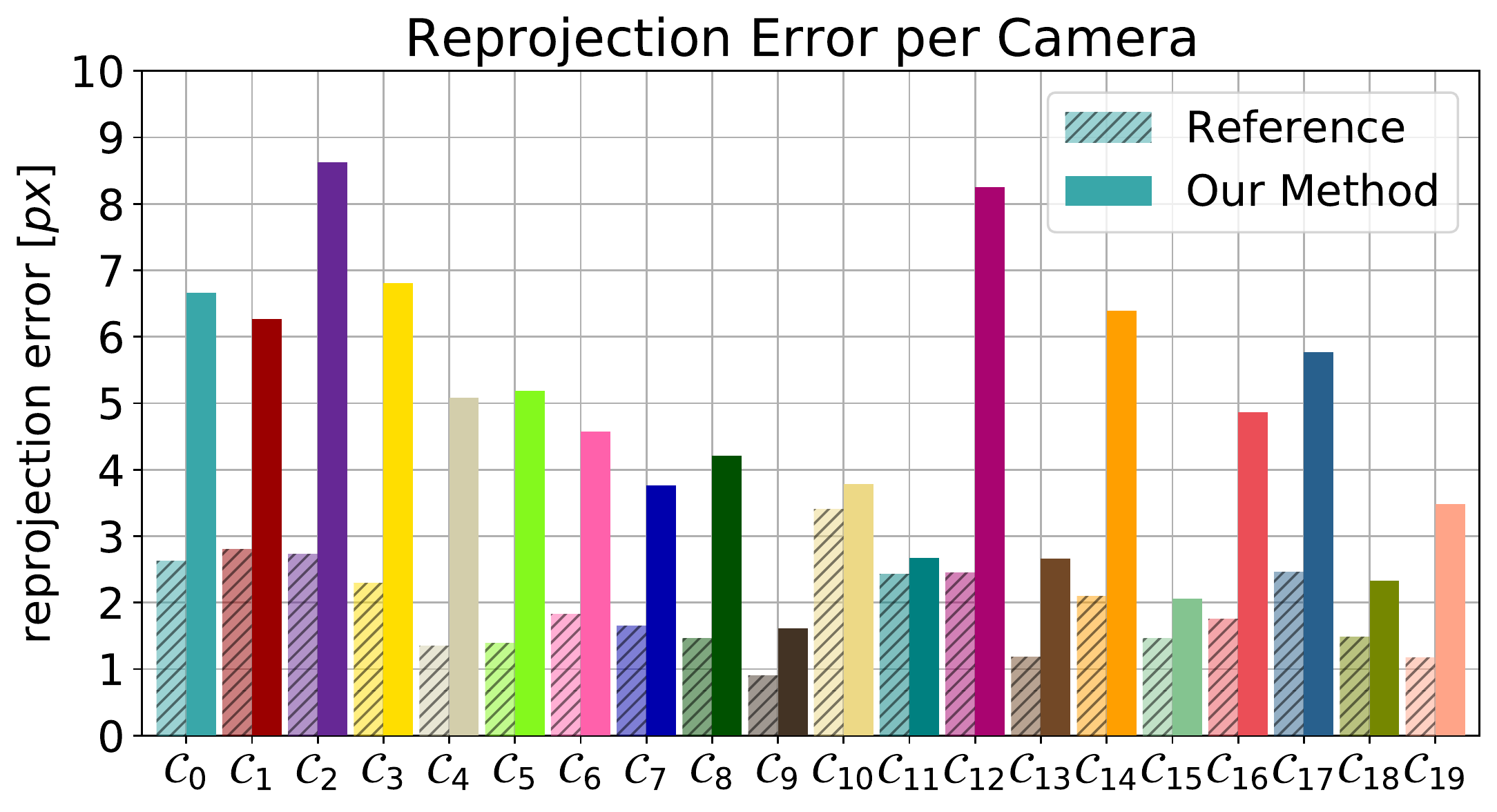}};

		\node[label,scale=1, anchor=south west, rectangle, fill=white, align=center, font=\scriptsize\sffamily] (n_0) at (image1.south west) {(a)};
		\node[label,scale=1, anchor=south west, rectangle, fill=white, align=center, font=\scriptsize\sffamily] (n_1) at (image2.south west) {(b)};
	\end{tikzpicture}				
	\vspace{-1em}
	\caption{Comparison of the reprojection error per camera between our method and the reference calibration using (a) keypoint- and (b) marker-based evaluation pipelines.}
	\label{fig:experiment4_3}
	\vspace{-1.5em}
\end{figure}
\reffig{fig:experiment4_3} shows the reprojection error per camera for both evaluation pipelines and \reftab{tab:mean_reprojection_per_cam} reports the averaged reprojection error. %
For the keypoint-based evaluation, we observe that our calibration achieves lower reprojection errors for all but two cameras.
For the marker-based evaluation, our calibration achieves similar reprojection errors as for the keypoint-based pipeline, while the reprojection errors of the reference calibration are significantly lower.
Our flexible, marker-free method achieves lower reprojection errors for the envisaged application of 3D multi-person pose estimation and still achieves a coherent result when evaluating with a traditional calibration target.
The difference in accuracy for the second evaluation is mainly due to our method being marker-free using features from persons of unknown dimensions for calibration, while the reference method knows the exact scale of the calibration (and evaluation) target.
Also, the noise in the joint detections may be larger than for the tag detections.

The averaged reprojection error per joint group for the keypoint-based evaluation is shown in \reftab{tab:experiment_4}. Our method achieves lower reprojection errors in all categories.
The reprojection error is larger for faster-moving joints like ankles and wrists, while it is smaller for more stable joints. 
This can be explained by limitations in the synchronization within framesets.

It is worth noting that the measured reprojection error does not exclusively originate from the provided extrinsic calibration, but also from other factors, e.g. the intrinsic camera calibration, or the approach for detection, data association, and triangulation.

\begin{table}[t]
	\centering
	\caption{Comparison of the average reprojection error of our method and the reference calibration for keypoint- and AprilTag-based evaluations, averaged over $10$ repetitions.}
	\vspace{-0.5em}
	\setlength{\tabcolsep}{3pt}
	\begin{tabular}{ |c|c|c| }
		\hline
		{\textbf{Calibration}}&
		{\textbf{Keypoints}}&
		{\textbf{AprilTag Grid}}\\[0.05cm]
		\hline\hline
		{Reference}&
		{\SI{4.57}{\pixel}}& %
		{\textbf{\SI{1.95}{\pixel}}}\\   
		\hline
		{Our Method}& 
		{\textbf{\SI{4.01}{\pixel}}}& %
		{\SI{5.00}{\pixel}}\\			
		\cline{0-2}      	
	\end{tabular}
	\label{tab:mean_reprojection_per_cam}
\end{table}

\begin{table}[t]
	\vspace{-0.3em}
	\centering
	\caption{Comparison of the reprojection errors per joint group between our method and the reference calibration averaged over $10$ repetitions of the experiment.}
	\vspace{-0.5em}
	\setlength{\tabcolsep}{3pt}
	\begin{tabular}{ |c|c|c|c|c|c|c|c| }		
		\hline
		{\textbf{Calibration}}&
		{\scriptsize \textbf{Head}}&            
		{\scriptsize \textbf{Hips}}&
		{\scriptsize \textbf{Knees}}&
		{\scriptsize \textbf{Ankles}}&
		{\scriptsize \textbf{Shlds}}&
		{\scriptsize \textbf{Elbows}}&
		{\scriptsize \textbf{Wrists}}\\[0.05cm]
		\hline\hline
		{\footnotesize Reference}& 
		{\footnotesize \SI{4.02}{\pixel}}&
		{\footnotesize \SI{5.10}{\pixel}}&
		{\footnotesize \SI{4.83}{\pixel}}&
		{\footnotesize \SI{5.88}{\pixel}}&
		{\footnotesize \SI{3.61}{\pixel}}&
		{\footnotesize \SI{4.29}{\pixel}}&
		{\footnotesize \SI{5.14}{\pixel}}\\
		\hline
		{\footnotesize Our Method}&
		{\footnotesize \textbf{\SI{3.55}{\pixel}}}&
		{\footnotesize \textbf{\SI{4.28}{\pixel}}}&
		{\footnotesize \textbf{\SI{4.20}{\pixel}}}&
		{\footnotesize \textbf{\SI{5.27}{\pixel}}}&
		{\footnotesize \textbf{\SI{3.21}{\pixel}}}&
		{\footnotesize \textbf{\SI{3.75}{\pixel}}}&
		{\footnotesize \textbf{\SI{4.55}{\pixel}}}\\
		\cline{0-7}      	
	\end{tabular}
	\label{tab:experiment_4}
\end{table}

\section{Conclusion}
In this work, we developed a marker-free online method for extrinsic camera calibration in a scene observed by multiple smart edge sensors, relying solely on person keypoint detections.
The keypoint detections are fused into 3D person hypotheses at a central backend by synchronization, filtering, and data association. Factor graph optimization problems are repeatedly solved to estimate the camera poses constrained by the observations. %
Knowledge about the camera poses obtained through the optimization of one factor graph is used during the construction of the next factor graph, enabling the accumulation of knowledge and the convergence of all cameras towards an accurate pose. %
Lastly, the convergence behavior is improved by a refinement scheme based on a Kalman filter. 

Our method is designed to be robust against false or sparse sets of detections and occlusions, and is free of many typical assumptions of similar methods:
It does not require a specific calibration target, can cope with and exploit the detections of multiple persons simultaneously, and handles arbitrary person poses. %
We evaluate the proposed method in a series of experiments and compare our calibration results to a reference calibration obtained by an offline calibration method based on traditional calibration targets.
We show that our calibration results are more accurate than the reference calibration by reliably achieving lower reprojection errors in a 3D multi-person pose estimation pipeline used as application scenario. Not only provides our method a quick and easy-to-use calibration utility, but it also achieves state-of-the-art accuracy.

The limitations of our method are mainly related to scaling ambiguity and data association:
The scale of the initial estimate of the extrinsic calibration is maintained throughout the calibration procedure. It inherently cannot resolve the scale in case the initial estimate is biased or inaccurate, as the dimensions of the persons used as calibration targets are unknown.
Data association could be improved for inaccurate initial estimates of the extrinsic calibration by using visual re-id descriptors.
We assume the intrisic camera calibration to be known. Our method could be extended by including intrinsic camera parameters in the optimization.
Finally, our method could be extended by also using additional environment features for calibration.

\bibliographystyle{splncs04}
\bibliography{References}

\end{document}